\documentclass[sigconf]{acmart}
\usepackage{balance}
\usepackage{multirow}
\usepackage{enumerate}
\usepackage{pifont}

\usepackage{enumitem}

\newcommand{\stitle}[1]{\vspace*{0.2em}\noindent{\bf #1\/}}
\newcommand{\squishlist}{
	\begin{list}{$\bullet$}
		{ \setlength{\itemsep}{1pt}
			\setlength{\parsep}{1pt}
			\setlength{\topsep}{2.5pt}
			\setlength{\partopsep}{0.5pt}
			\setlength{\leftmargin}{1em}
			\setlength{\labelwidth}{1em}
			\setlength{\labelsep}{0.6em}
		}
	}
	\newcommand{\squishend}{
	\end{list}
}

\AtBeginDocument{%
  }

\setcopyright{acmlicensed}
\copyrightyear{2018}
\acmYear{2018}
\acmDOI{XXXXXXX.XXXXXXX}

\acmConference[Conference acronym 'XX]{Make sure to enter the correct
  conference title from your rights confirmation emai}{June 03--05,
  2018}{Woodstock, NY}
\acmISBN{978-1-4503-XXXX-X/18/06}

\begin{document}
% \begin{sloppypar}

\title{Grid and Road Expressions Are Complementary for Trajectory Representation Learning}

\author{Silin Zhou}
\affiliation{%
  \institution{University of Electronic Science and Technology of China}
  \city{Chengdu}
  \country{China}
}
\email{zhousilinxy@gmail.com}

\author{Shuo Shang}
\authornote{The corresponding author.}
\affiliation{%
  \institution{University of Electronic Science and Technology of China}
  \city{Chengdu}
  \country{China}
}
\email{jedi.shang@gmail.com}

\author{Lisi Chen}
\affiliation{%
  \institution{University of Electronic Science and Technology of China}
  \city{Chengdu}
  \country{China}
}
\email{lchen012@e.ntu.edu.sg}

\author{Peng Han}
\affiliation{%
  \institution{University of Electronic Science and Technology of China}
  \city{Chengdu}
  \country{China}
}
\email{penghan_study@foxmail.com}

\author{Christian S. Jensen}
\affiliation{%
 \institution{Aalborg University}
 \city{Aalborg}
 \country{Denmark}
}
\email{csj@cs.aau.dk}


\begin{abstract}
Trajectory representation learning (TRL) maps trajectories to vectors that can be used for many downstream tasks. Existing TRL methods use either \textit{grid trajectories}, capturing movement in free space, or \textit{road trajectories}, capturing movement in a road network, as input. We observe that the two types of trajectories are complementary, providing either region and location information or providing road structure and movement regularity. Therefore, we propose a novel multimodal TRL method, dubbed \textbf{GREEN}, to jointly utilize \textbf{G}rid and \textbf{R}oad trajectory \textbf{E}xpressions for \textbf{E}ffective representatio\textbf{N} learning. In particular, we transform raw GPS trajectories into both grid and road trajectories and tailor \textit{two encoders} to capture their respective information. 
To align the two encoders such that they complement each other, we adopt a \textit{contrastive loss} to encourage them to produce similar embeddings for the same raw trajectory and design a \textit{mask language model} (MLM) loss to use grid trajectories to help reconstruct masked road trajectories. To learn the final trajectory representation, a \textit{dual-modal interactor} is used to fuse the outputs of the two encoders via cross-attention. 
We compare GREEN with 7 state-of-the-art TRL methods for 3 downstream tasks, finding that GREEN consistently outperforms all baselines and improves the accuracy of the best-performing baseline by an average of 15.99\%. Our code and data are available online\footnote{https://github.com/slzhou-xy/GREEN}.
\end{abstract}



\maketitle

\section{Introduction} \label{section-intro}
Trajectories record the movements of objects (e.g., vehicles or pedestrians).
Analyzing trajectory data is crucial for understanding movement patterns and helps to solve many real-world problems, including transportation optimization~\cite{transportation_optimization}, traffic management~\cite{congestion_management}, and human mobility analysis~\cite{human_mobility}. These problems rely on a set of basic trajectory operations such as trajectory classification~\cite{TrajODE}, travel time estimation~\cite{iETA}, and trajectory similarity computation~\cite{gts}. 

\textit{Trajectory representation learning} (TRL) is the task of learning vector representations of variable-length trajectories to facilitate these operations~\cite{trajectory_survey}. 
For example, to compute trajectory similarity, traditional methods have a quadratic cost w.r.t. trajectory length~\cite{sim1,sim2,edr} as they adopt dynamic programming, instead using fixed-length vectors is much more efficient.

\begin{figure}[!t]
    \centering
    \includegraphics[width=0.90\linewidth]{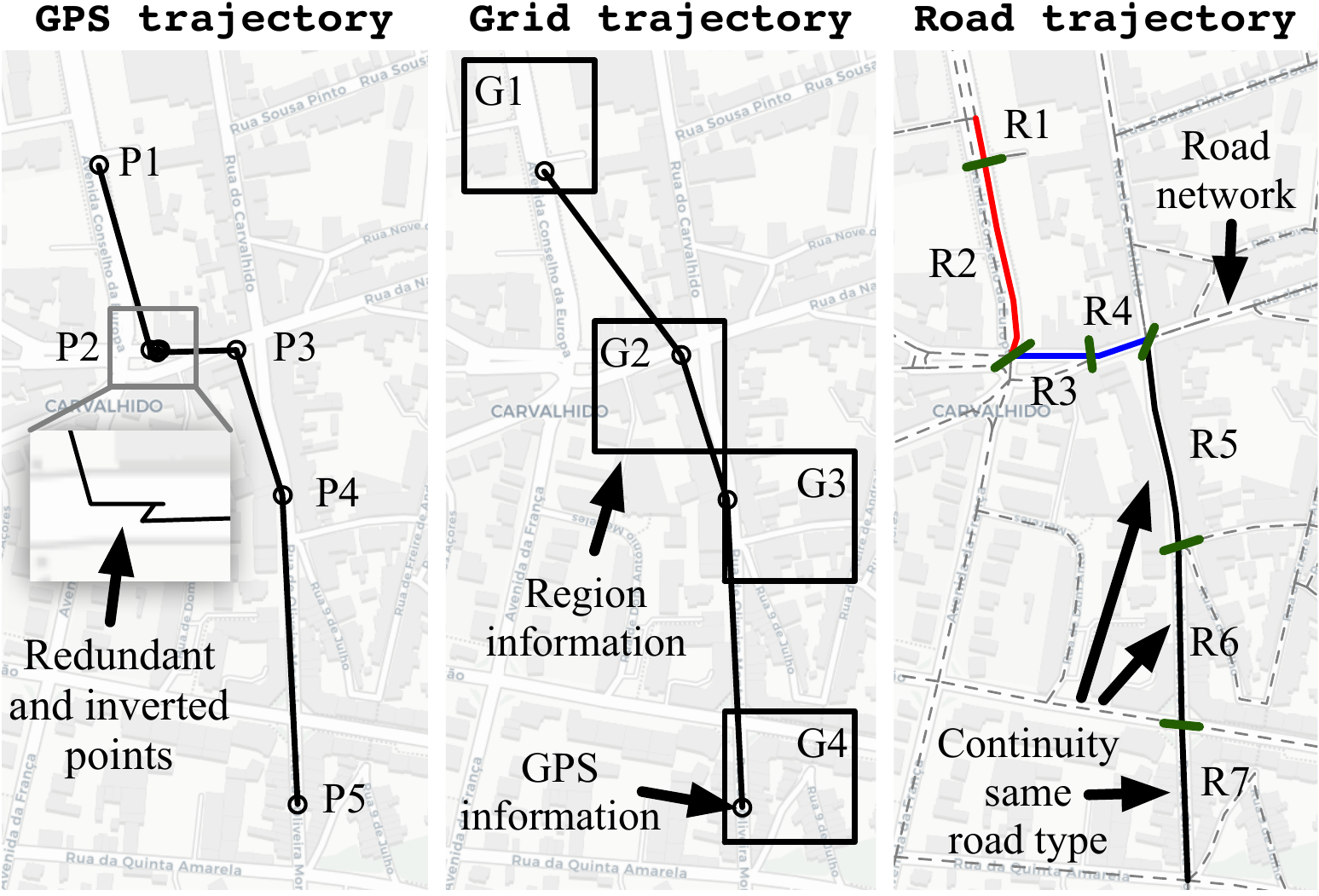}
    \caption{Comparison of three types of trajectories.}
    \label{fig:intro}
\end{figure}

A trajectory can be expressed in three different forms: as a \textit{GPS trajectory}, a \textit{grid trajectory}, or a \textit{road trajectory}, as shown in Figure~\ref{fig:intro}. In particular, a GPS trajectory is a sequence of GPS points sampled at regular time intervals, with the points being timestamped coordinates. A GPS trajectory is often noisy due to positioning inaccuracies (e.g., P2 in Figure~\ref{fig:intro}), and this noise is harmful to downstream processing. Thus, only early TRL methods (e.g., Traj2vec~\cite{traj2vec} and CSTRM~\cite{CSTRM}) use raw GPS trajectories as inputs. Grid trajectories are obtained by covering the space with a grid and mapping GPS points to the grid cell that contains them. Thus, a grid trajectory is a sequence of timestamped grid cells. Compared with a GPS trajectory, a grid trajectory preserves location and time information but is more robust to noise as multiple GPS points can be mapped to the same cell. Moreover, grid trajectories can also encode information about urban regions (e.g., downtown or residential) since a grid cell covers a fixed region so that properties such as trajectory density and average movement speed can be associated with such regions. Therefore, methods like t2vec~\cite{t2vec}, E$^2$DTC~\cite{e2dtc}, and TrajCL~\cite{trajcl} use grid trajectories. Next road trajectories are obtained by map matching~\cite{st_matching,fmm}) GPS trajectories to an underlying road network, so that road trajectories are timestamped sequences of road segments. Like grid trajectories, road trajectories can reduce noise due to the map matching. By incorporating a road network, road trajectories model movement regularities, as vehicles can only travel along the road segments. The majority of recent TRL methods, such as START~\cite{start}, JGRM~\cite{jgrm}, and JCLRNT~\cite{JCLRNT}, use road trajectories.

We observe that existing TRL methods only use a single type of trajectory representation, although different trajectory expressions capture complementary information. Specifically, grid trajectories reduce noise in raw GPS points and excel at capturing location and time information. They also capture information about urban regions. In contrast, road trajectories capture movement regularity along road segments. These trajectories also capture information about the road network. For example, deriving properties, traffic volumes, travel speeds of road segments, and transition probabilities between road segments. Such information can be used to enhance the trajectory representation in turn. 

We propose GREEN to jointly utilize grid and road trajectory expressions for TRL. By exploiting both expressions, GREEN aims to improve over existing TRL methods. GREEN resembles multi-modal learning~\cite{clip,albef,flip}, as it treats the two trajectory expressions as different modalities. Like multi-modal learning, GREEN addresses two challenges to make its idea work.

\textit{\ding{182} How to effectively capture the information in grid and road trajectory expressions?} GREEN includes an encoder for each trajectory expression to capture its particular information. Thus, the grid encoder uses a convolution neural network (CNN)~\cite{cnn} and treats the grid cells as pixels in an image and uses the traffic flow and geographical locations of the cells as pixel values. To encode more information, we augment the cell representations learned by the CNN with spatial-temporal information unique to each GPS trajectory. The road encoder uses a graph neural network (GNN)~\cite{gnn} to learn representations with structure information from the road segments in the road network graph. We also enhance the road representations with road type and trajectory-specific timings to capture more information. Both the grid encoder and road encoder adopt Transformer~\cite{transformer} as the backbone by modeling trajectories as sequences of grid and road representations. The outputs of the two encoders are the grid and road representations of trajectories.  

\textit{\ding{183} How to integrate the grid and road trajectory representations?} We need to first ensure that the grid and road representations align with each other and then fuse them into a final trajectory representation. We design two loss functions for alignment, i.e., contrastive loss~\cite{clip} and masked language model (MLM) loss~\cite{bert}. The contrastive loss encourages the grid and road encoder to produce similar representations for the same trajectory such that the representations belong to the same space and can be fused. The MLM loss reconstructs a complete trajectory from a masked road trajectory and uses the grid representation to facilitate the reconstruction. As such, it encourages the grid representation to complement the road representation. Since the contrastive loss uses complete road trajectories while the MLM loss uses masked road trajectories, we perform two forward operations over the road encoder with different inputs. Finally, we use an interactor to fuse the grid and road representations via cross-attention. 

To evaluate GREEN, we compare with 7 state-of-the-art TRL methods and experiment on 2 real-world datasets with 3 important downstream tasks: travel time estimation, trajectory classification, and most similar trajectory search. The results show that GREEN is able to consistently outperform all baselines in terms of accuracy across all tasks and datasets. In particular, compared with the best-performing baseline, the average accuracy improvement of GREEN for travel time estimation, trajectory classification, and most similar trajectory search are 19.55\%, 2.21\%, and 23.90\%, respectively. An ablation study shows that all elements of GREEN contribute to improving accuracy. We also observe that GREEN is comparable to the baselines in terms of training and inference efficiency. 

To summarize, we make the following contributions:
\squishlist
    \item We observe that existing TRL methods use only one trajectory expression and propose to jointly utilize grid and road trajectory expressions to capture more information for enhanced accuracy.  

    \item We design GREEN as a self-supervised learning model for TRL, featuring encoders tailored to extract information from both grid and road trajectory expressions, and we propose specialized loss terms to ensure that the two expressions complement each other.
     
    \item We evaluate GREEN experimentally, by comparing it with 7 state-of-the-art TRL methods and on 3 downstream tasks, validating our designs in ablation study, and quantifying its efficiency.
\squishend

\section{Related Work}
Existing solutions to the TRL problem can be classified into GPS-based TRL, grid-based TRL, and road-based TRL.

\stitle{GPS-based TRL.} The TRL problem can be addressed by learning representations from GPS trajectories in free space. For example, traj2vec~\cite{traj2vec} transforms a GPS trajectory to a trajectory feature sequence and applies an RNN-based seq2seq~\cite{seq2seq} model to learn trajectory representations.
CSTRM~\cite{CSTRM} learns trajectory representations by distinguishing trajectory-level and point-level differences between GPS trajectories. However, GPS points carry limited spatial-temporal information and are noisy~\cite{trajectory_survey}, e.g., due to drift points, inversion points, and redundant points, which can lead to inaccurate trajectory modeling. Thus, only early methods use GPS trajectories directly. Our study uses grid and road trajectories to capture more information and reduce noise in GPS trajectories.

\stitle{Grid-based TRL.} To simplify or enhance GPS trajectories in free space, some methods~\cite{t2vec,e2dtc,trajgat,trajcl} use grids to simplify GPS trajectories and learn grid trajectory representations. For example, t2vec~\cite{t2vec} first applies an encoder-decoder model~\cite{seq2seq} to learn robust grid trajectory representations from low-quality data to support trajectory similarity computation and search. E$^2$DTC aims to capture hidden spatial dependencies in trajectories and proposes self-training to learn grid trajectory representations for trajectory clustering. TrajCL~\cite{trajcl} introduces various trajectory augmentations and uses contrastive learning to jointly learn grid trajectory representations from spatial and structural perspectives. 
Unlike the above methods, 
TrajGAT~\cite{trajgat} abandons the sequential model in favor of a graph-based model to learn trajectory representations. 
In our study, we employ a CNN to capture regional properties of grid cells, while incorporating GPS features to improve spatial-temporal information capture and enhance accuracy of grid representations.

\begin{figure*}[!t]
    \centering
    \includegraphics[width=0.9\linewidth]{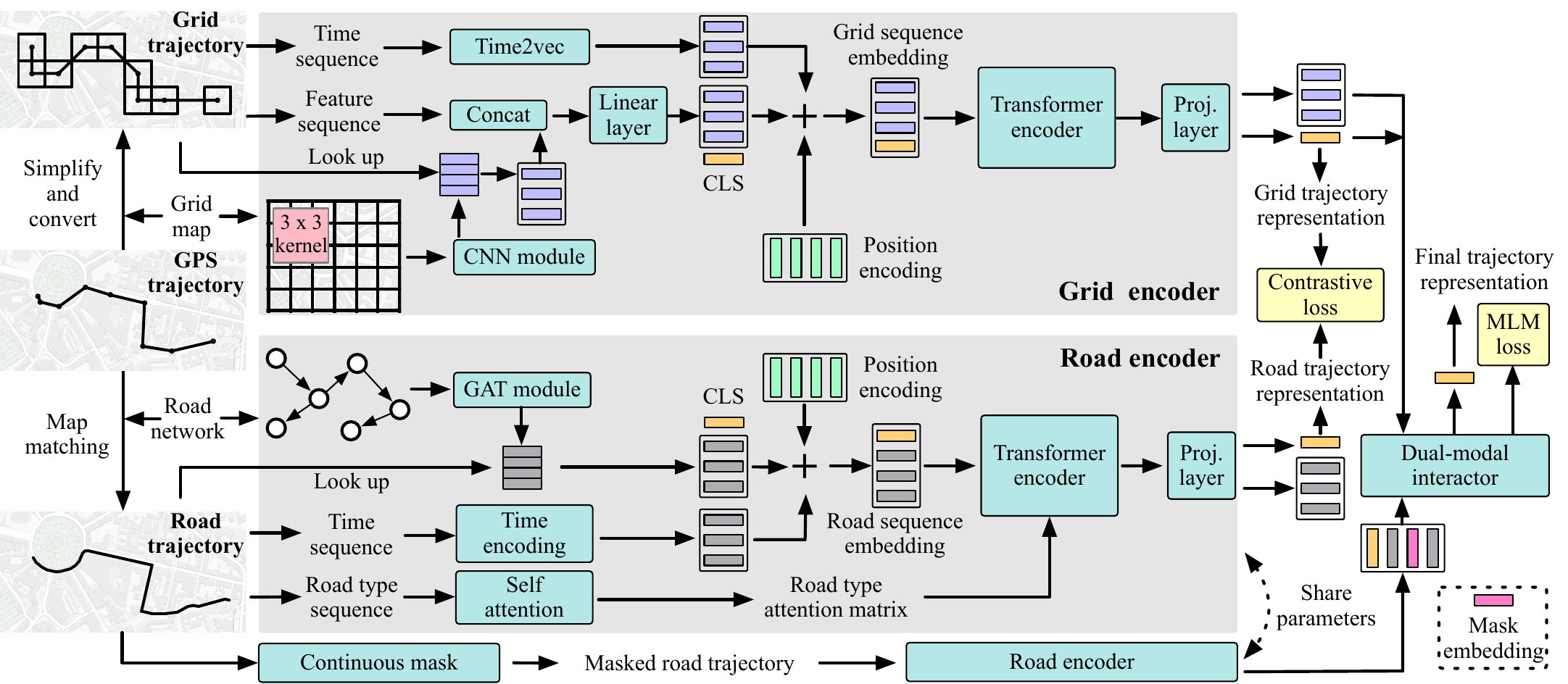}
    \caption{Overview of GREEN.}
    \label{fig:framework}
\end{figure*}

\stitle{Road-based TRL.}
Road trajectories, obtained using map matching~\cite{st_matching,fmm}, capture the transportation structure of a region~\cite{trajectory_survey}. Some studies learn road trajectory representations for specific tasks, e.g., similarity computation~\cite{gts,ST2Vec,grlstm}, trajectory anomalous detection~\cite{Anomalous_Trajectory_Detection}, and trajectory generation~\cite{RnTrajRec}. However, the representations obtained by these methods are difficult to transfer to other spatial-temporal tasks. To learn generic road trajectory representations, Trembr~\cite{trembr} uses an RNN-based seq2seq model~\cite{seq2seq} and uses spatial-temporal information of road segments to learn representations. PIM~\cite{pim} and Toast~\cite{toast} first use node2vec~\cite{node2vec} to learn road embeddings on the road network. Then PIM utilizes mutual information maximization, and Toast introduces route recovery tasks to reconstruct masked road segments to learn trajectory representations, respectively. 
JCLRNT~\cite{JCLRNT} creates specialized augmentations for road-road contrast using contextual neighbors and for trajectory-trajectory contrast by replacing detours and removing alternatives. 
START~\cite{start} integrates travel semantics and temporal regularities into the BERT~\cite{bert}, utilizing various data augmentations such as trimming and temporal shifting of trajectories to enhance the accuracy and robustness of trajectory representations. JGRM~\cite{jgrm} enriches the spatial-temporal information of roads and models road trajectories by aligning GPS points to corresponding road segments. Our study considers the information of grid and road trajectories, and we integrate them to obtain the final representation.

\section{Preliminaries}
\label{sec:problem_formulation}
\subsection{Definitions}
\noindent{\bf Definition 1 (GPS Trajectory).} A GPS trajectory $T^{gps}$ is a sequence of points collected at a fixed sampling time interval by a GPS-enabled device. Each point is of the form $p_i = (x_i, y_i, t_i)$, where $x_i$, $y_i$, and $t_i$ denote longitude, latitude, and timestamp, respectively.

\stitle{Definition 2 (Grid Trajectory).} A grid trajectory $T^{g}$ is a sequence of timestamped grid cells obtained by transforming GPS points in $T^{gps}$. Each cell is of the form $\tau_i^g = (g_i, t_i^g)$, where $g_i$ denotes the ID of a grid and $t_i^g$ denotes the time.

\stitle{Definition 3 (Road Network).} A road network is modeled as a directed graph $G = ( V, A )$, where $ V $ denotes the road segment set in the road network, and $A \in \mathbb{R}^{\vert V \vert \times \vert V \vert}$ is the adjacency matrix that represents the connectivity between road segments. $A[i, j] = 1$ if and only if road segments $v_i$ and $v_j$ are directly connected, otherwise $A[i, j] = 0$. Under this definition, a trajectory can be extracted as a sequence of road segments that it passes through.

\stitle{Definition 4 (Road Trajectory).} A road trajectory $T^{r}$ is a timestamped sequence of road segments that is generated from GPS trajectory $T^{gps}$ using map matching. Each road segment is of the form $\tau_i^r = (v_i, t_i^r)$, where $v_i$ denotes the ID of a road segment in the road network $G$ and $t_i^r$ denotes the time.

\subsection{Problem Statement}\label{subsec:problem}
We are given a set of GPS trajectories and a road network. Then we aim to learn a $d$-dimensional vector representation of each trajectory in the set by using the two expressions, i.e., grid trajectories and road trajectories. We expect the learned representation to achieve high accuracy for downstream tasks such as:
% \squishlist
\begin{itemize}[leftmargin=*]
    \item \textbf{Travel time estimation}: Given a trajectory $T_a$ without time information, this task predicts the travel time of $T_a$. 
    \item \textbf{Trajectory classification}: Given a trajectory $T_a$, this task assigns $T_a$ to a category, e.g., the type of vehicle. 
    \item \textbf{Most similar trajectory search}: Given a query trajectory $T_a$ and a trajectory database $\mathcal{D}$, this task finds the trajectory $T_b \in \mathcal{D}$ that is most similar to $T_a$.
\end{itemize}
% \squishend

\section{The GREEN Model} \label{section:method}
We proceed to present GREEN. Figure~\ref{fig:framework} provides an overview of GREEN, which consists of three modules, i.e., a \textit{grid encoder} (Section~\ref{section:grid_encoder}), a \textit{road encoder} (Section~\ref{section:road_encoder}), and a \textit{dual-modal interactor} (Section~\ref{section:interactor}). We use two self-supervised tasks (Section~\ref{section:loss}) to train GREEN, i.e., \textit{contrastive loss} and \textit{MLM loss}. For conciseness, we ignore multi-head attention and the [CLS] token in formulas related to the Transformer architecture.

\subsection{Grid Encoder} \label{section:grid_encoder}
The grid encoder is designed to learn grid trajectory representations in free space. To obtain grid trajectories, the underlying space is first partitioned using a regular grid, and then GPS points are mapped to the ID of the grid cells they fall into. Compared to GPS trajectories, grid trajectories retain location information and capture rich regional information. We design a grid encoder to learn region and spatial-temporal information of grid trajectories.

\stitle{CNN-based Grid Learning.} We first learn regional information from a global view. In particular, a grid map of $H$-height and $W$-width can be transformed into a 2D-image $M \in \mathbb{R}^{H \times W \times C}$, where $(H, W)$ is the resolution of the image and $C$ is the number of channels. The channels represent the features of the grid cells, and we use three types of region information: $( \mathbf{x}_i, \mathbf{y}_i, \mathbf{c}_i )$. Thus, $C = 3$, and $( \mathbf{x}_i, \mathbf{y}_i)$ denotes the central coordinate of the grid, while $\mathbf{c}_i$ captures the traffic flow (the number of times trajectories in the training dataset visits a grid cell) of cell $g_i$. Then, we use a CNN to learn global region representations for the grids in the 2D-image:
\begin{equation}    
\begin{aligned}
    \mathbf{X}^g_1 &= \sigma(\textrm{Conv}_1(M)), \\
    \mathbf{X}^g_2 &= \sigma(\textrm{Conv}_2(\mathbf{X}^g_1)),
\end{aligned}
\end{equation}
where $\sigma$ is the relu~\cite{relu} activation function, 
% $\textrm{Bn}$ is the batch normalization~\cite{bn}, 
$\textrm{Conv}_1$ and $\textrm{Conv}_2$ are convolution kernels of size $3 \times 3$. Note that $\mathbf{X}^g_2 \in \mathbb{R}^{H \times W \times C^\prime}$, where $C^\prime$ is the number of output channels. To learn grid embeddings with hidden dimension $h$, we reshape $\mathbf{X}^g_2$ into a 2D-tensor, and use a 2-layer MLP to map it:
\begin{equation}
    \mathbf{X}^g = \textrm{MLP}(\textrm{Reshape}(\mathbf{X}^g_2)),
\end{equation}
where $\mathbf{X}^g \in \mathbb{R}^{(H \cdot W) \times h}$. We take $\mathbf{X}^g$ as the grid embedding table and transform each grid ID into an embedding by means of a lookup in $\mathbf{X}^g$. Thus, we can transform a grid trajectory $T^g = \langle \tau^g_1, \tau^g_2, ..., \tau^g_{\vert T^g \vert} \rangle$ into a grid embedding sequence $\mathbf{E}^g = \langle \mathbf{e}^g_1, \mathbf{e}^g_2, ..., \mathbf{e}^g_{\vert T^g \vert} \rangle$.

\stitle{Feature-enhanced Grid Representation.} Although the grid embedding contains rich region information, it is the same for different trajectories. However, the same grid ID has different semantics for different trajectories. To address this issue, we incorporate GPS information for a grid trajectory $T^g$. For each grid $\tau^g_i = (g_i, t_i^g) \in T^g$, we choose a GPS point as the anchor, and if there are multiple consecutive GPS points of a trajectory in $g_i$, we choose the first GPS point (The statistical results show that less than 1\% of the grids under 100m$\times$100m in the dataset have more than one GPS point). For the spatial information, we compute a 4-dimensional vector $\hat{\mathbf{e}}_i^g = (\mathbf{x}_i^g, \mathbf{y}_i^g, \mathbf{d}_i^g, \mathbf{r}_i^g)$
for each grid $\tau^g_i \in T^g$, where $\mathbf{x}_i^g$ and $\mathbf{y}_i^g$ are GPS coordinates, $\mathbf{d}_i^g$ is the distance to the previous GPS point and $\mathbf{r}_i^g$ is the azimuth w.r.t. the previous GPS point. Then we use a linear layer to fuse the general $\mathbf{e}_i^g$ and the trajectory specific $\hat{\mathbf{e}}_i^g$:
\begin{equation}
    \tilde{\mathbf{e}}_i^g = \textrm{Linear}( \mathbf{e}_i^g \parallel \hat{\mathbf{e}}_i^g ),
\end{equation}
where $\parallel$ denotes vector concatenation. For the temporal information, we use pre-trained Time2vec~\cite{time2vec} to encode time embedding for the fine-grained time $t_i^g$, i.e., timestamp, as follows:
\begin{equation}
    \mathbf{t}_i = f_1(t_i^g) \parallel \text{sin}(f_2(t_i^g)),
\end{equation}
where $f_1(\cdot)$ and $f_2(\cdot)$ are linear layers to learn time embedding, and $\textrm{sin}(\cdot)$ function helps capture time periodic behaviors. Then we use a Transformer to learn the grid trajectory representation as follows:
\begin{equation}    
\begin{aligned}
    \mathbf{h}_i^g &= \tilde{\mathbf{e}}_i^g + \mathbf{t}_i^g + \mathbf{p}_i^g, \\
    \mathbf{H}^g &= [\mathbf{h}_1^g,  \mathbf{h}_2^g, ..., \mathbf{h}_{\vert T^g \vert}^g ] \in \mathbb{R}^{\vert T^g \vert \times h}, \\
    \mathbf{Z}_1^g &= \textrm{TransformerEncoder}(\mathbf{H}^g), \\
\end{aligned}
\end{equation}
where $\mathbf{p}_i^g$ is the position encoding of grid $\tau_i^g$, $\mathbf{h}_i^g$ is the hidden embedding of the Transformer, and $\mathbf{Z}_1^g \in \mathbb{R}^{\vert T^g \vert \times h}$. Next, we use a linear layer to map $\mathbf{Z}_1^g$ into a new embedding space with dimension $d$:
\begin{equation}
    \mathbf{Z}_2^g = \textrm{Linear}(\mathbf{Z}_1^g) \in \mathbb{R}^{\vert T^g \vert \times d}.
\end{equation}
The final grid trajectory representation $\mathbf{v}^g$ is obtained by the embedding of the [CLS] token into $\mathbf{Z}_2^g$.

\subsection{Road Encoder} \label{section:road_encoder}
The road encoder aims to learn road trajectory representations on the road network. A road trajectory, which is obtained by map matching, captures the actual route of the object. Compared with a GPS trajectory, a road trajectory is continuous and includes road segment transfer information. The road network also introduces information on urban transportation structures. We design a road encoder to learn urban structure and continuity of road trajectories.

\stitle{GNN-based Road Structure Learning.} We first capture the topological attributes of the road network using a graph neural network (GNN) at the global level. 
In particular, we concatenate multiple attributes\footnote{These attributes are provided as part of the downloaded road network data.} as features of road segments, including maximum speed limit, average travel time, road direction, out-degree, in-degree, road length, and
road type (eight class for one-hot encoding, \{\textit{living street}, \textit{motorway}, \textit{primary}, \textit{residential}, \textit{secondary}, \textit{tertiary}, \textit{trunk}, \textit{unclassified}\}), which is denoted as $F \in \mathbb{R}^{\vert V \vert \times 14}$. We learn the road embeddings with topological structure as follows:
\begin{equation}    
\begin{aligned}
    \mathbf{X}^r_1 &= \textrm{Linear}(F), \\
    \mathbf{X}^r_2 &= \textrm{GAT}(G, \mathbf{X}^r_1),
\end{aligned}
\end{equation}
where $\mathbf{X}^r_1, \mathbf{X}^r_2 \in \mathbb{R}^{\vert V \vert \times h}$, and $\textrm{GAT}$ is the graph attention network~\cite{gat}. Like the grid trajectory, we can transform a road trajectory $T^r = \langle \tau^r_1, \tau^r_2, ..., \tau^r_{\vert T^r \vert} \rangle$ into a road embedding sequence $\mathbf{E}^r = \langle \mathbf{e}^r_1, \mathbf{e}^r_2, ..., \\ \mathbf{e}^r_{\vert T^r \vert} \rangle$ by looking up road embeddings in $\mathbf{X}^r_2$ using road IDs.

\stitle{Continuity-enhanced Road Representation.} We observe that some road segments are aligned many GPS points with many timestamps or are generated completely by the map matching with no corresponding GPS points, which makes it difficult to assign a precisely timestamp to them (The statistical results show that more than 20\% of the road segments in the dataset have multiple GPS points or are completed by map matching). To solve this problem, we use a coarse-grained time encoding scheme. For roads with many GPS points or without GPS points, the number of minutes can be estimated accurately based on the context. Therefore, we use two learnable matrices for time encoding on road segments, $\mathbf{E}^{day} \in \mathbb{R}^{1440 \times h}$ to represent the minute of the day and $\mathbf{E}^{week} \in \mathbb{R}^{7 \times h}$ to represent the day of the week. This approach solves the road temporal misalignment problem and also introduces the second modal information into the temporal encoding.
Thus, the input embedding to the Transformer of a road trajectory $T^r$ is as follows:
\begin{equation}    
\begin{aligned}
    \mathbf{h}_i^r &= \mathbf{e}_i^r + \mathbf{e}_i^{day} + \mathbf{e}_i^{week} + \mathbf{p}_i^r, \\
    \mathbf{H}^r &= [\mathbf{h}_1^r,  \mathbf{h}_2^r, ..., \mathbf{h}_{\vert T^r \vert}^r ] \in \mathbb{R}^{\vert T^r \vert \times h}, 
\end{aligned}
\end{equation}
where $\mathbf{e}_i^{day} \in \mathbf{E}^{day}$ and $ \mathbf{e}_i^{week} \in \mathbf{E}^{week}$ are time embeddings corresponding to $t_i^r$ in $\tau_i^r \in T^r$, and $\mathbf{p}_i^r$ is the position encoding. 

However, the vanilla Transformer cannot capture the local continuity of road trajectories. When vehicles are traveling in a road network, the types of adjacent road segments are usually the same\footnote{The road types stem from the downloaded road network data and are divided into eight categories, i.e., \{living street, motorway, primary, residential, secondary, tertiary, trunk, unclassified\}.}, as shown in Figure~\ref{fig:intro}. To model such continuity, we explicitly inject road type information to dynamically adjust the self-attention coefficient. We first use a learnable road type matrix $\mathbf{E}^{\mathit{type}} \in \mathbb{R}^{\vert \mathit{type} \vert \times h}$ to transform the road type sequence $O = \langle o_1, o_2, ..., o_{\vert T^r \vert} \rangle$ of $T^r$ into a road type embedding $\mathbf{E}^{o} = \langle e_1^o, e_2^o, ..., e^o_{\vert T^r \vert} \rangle$, and then we incorporate road type information into the Transformer as follows:
\begin{equation}
\begin{aligned}
    \mathbf{H}^o &= [ \mathbf{h}_1^o,  \mathbf{h}_2^o, ..., \mathbf{h}_{\vert T^r \vert}^o ] \in \mathbb{R}^{\vert T^r \vert \times h}, \mathbf{h}_i^o         = \mathbf{e}_i^o + \mathbf{p}_i^o, \\
    \mathbf{A}^o &= \frac{\mathbf{Q}^o \mathbf{K}^{o^\top}}{\sqrt{h}}, \mathbf{Q}^o = \mathbf{H}^o W^o_q, \mathbf{K}^o = \mathbf{H}^o W^o_k, \\
    \mathbf{A}^r &= \frac{\mathbf{Q}^r \mathbf{K}^{r^\top}}{\sqrt{h}}, \mathbf{Q}^r = \mathbf{H}^r W^r_q, \mathbf{K}^r = \mathbf{H}^r W^r_k, \mathbf{V}^r = \mathbf{H}^r W^r_v,  \\
    \mathbf{Z}^r_1 &= \textrm{FFN}\big( \textrm{Softmax}(\mathbf{A}^r + \mathbf{A}^o) \mathbf{V}^r \big),
\end{aligned}
\end{equation}
where $\mathbf{p}_i^o$ is the position encoding, $W^o_q$, $W^o_k$, $W^r_q$, $W^r_k$, $W^r_v \in \mathbb{R}^{h \times h}$ are learnable matrices, $\textrm{FFN}$ is the feed-forward network corresponding to the vanilla Transformer, and $\mathbf{Z}_1^r \in \mathbb{R}^{\vert T^r \vert \times h}$. Then, we map $\mathbf{Z}_1^r$ into a new embedding space with dimension $d$:
\begin{equation}
    \mathbf{Z}^r_2 = \textrm{Linear}(\mathbf{Z}^r_1) \in \mathbb{R}^{\vert T^r \vert \times d}.
\end{equation} 
The final road trajectory representation $\mathbf{v}^r$ is obtained by the embedding of the [CLS] token into $\mathbf{Z}_2^r$.

\subsection{Dual-modal Interactor} \label{section:interactor}
The dual-modal interactor fuses grid and road representations into the final trajectory representation. 
Grid and road trajectories are two expressions of GPS trajectories in geographic free space and road network space, respectively, each containing unique spatial-temporal information, e.g., region and location information for grid trajectory, road network, and movement continuity for road trajectory. 
To enable the final trajectory representation to learn from the grid and road expressions of trajectories, we use cross-attention as the dual-modal interactor as follows:
\begin{equation}
\begin{aligned}
     \mathbf{Q}^m &= \mathbf{Z}_2^r W^m_q, \mathbf{K}^m = \mathbf{Z}_2^g W^m_k, \mathbf{V}^m = \mathbf{Z}_2^g W^m_v,  \\
    \mathbf{Z}^m &= \textrm{FFN}\big( \textrm{Softmax}(\frac{\mathbf{Q}^m \mathbf{K}^{m^\top}}{\sqrt{d}}) \mathbf{V}^m \big),
\end{aligned}
\end{equation}
where $W^m_q$, $W^m_k$, $W^m_v \in \mathbb{R}^{d \times d}$ are learnable matrices, and $\mathbf{Z}^m \in \mathbf{R}^{\vert T^r \vert \times d}$. Note that the query matrix $\mathbf{Q}^m$ comes from road representation, while the key matrix $\mathbf{K}^m$ and value matrix $\mathbf{V}^m$ come from grid representations. This is because we use grid trajectories to help reconstruct masked road trajectories in the MLM loss during training, which is covered in Section~\ref{section:loss}. 
The final trajectory representation is the embedding of the [CLS] token into $\mathbf{Z}^m$.

\subsection{Loss Functions} \label{section:loss}
We use two self-supervised loss functions for training, i.e., contrastive loss and MLM loss. 

\stitle{Contrastive Loss.} The contrastive loss is applied to the grid and road representations such that they are aligned to prepare for the subsequent fusion by the dual-modal interactor. Given a similarity function $s(\cdot, \cdot)$ that measures how similar a grid trajectory representation and a road trajectory representation are, the contrastive loss encourages the grid and road encoders to produce similar representations for the same raw GPS trajectory and dissimilar representations for different raw GPS trajectories. 
Given a batch of training trajectories, we compute the softmax-normalized grid-to-road trajectory loss as follows:
\begin{equation}
\begin{aligned}
     s(T^g, T^r) &= \frac{\mathbf{v}^g \cdot \mathbf{v}^r}{\vert \mathbf{v}^g \vert \vert \mathbf{v}^r \vert}, \\
     \mathcal{L}^g = -\frac{1}{N} \sum_{i=1}^{N}  \log & \frac{\exp (s (T^g_i, T^r_i) / \delta)}{\sum_{j=1}^{N} \exp (s (T_i^g, T^r_j) / \delta)}, 
\end{aligned}
\end{equation}
where $N$ is the batch size and $\delta$ is a learnable temperature parameter. We compute a road-to-grid trajectory loss $\mathcal{L}^r$ in the same way, and the final contrastive loss in a batch is $\mathcal{L}^{cl} = \frac{\mathcal{L}^g + \mathcal{L}^r}{2}$.

\begin{figure}[!t]
    \centering
    \includegraphics[width=\linewidth]{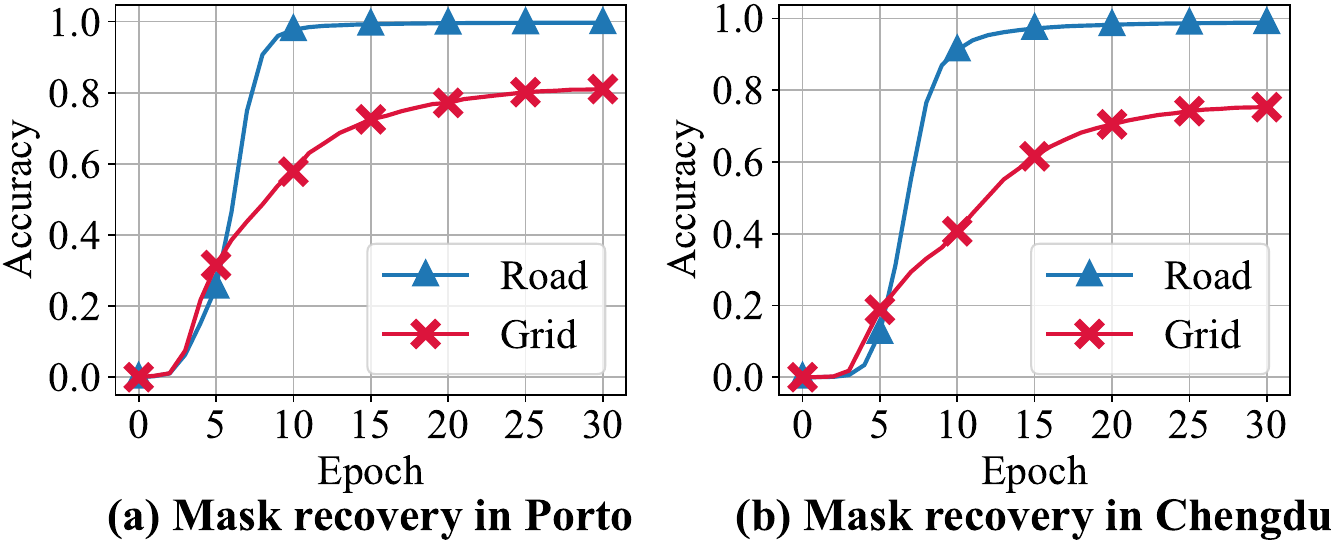}
    \caption{Comparison of grid and road mask recovery.}
    \label{fig:mask_recover}
\end{figure}

\stitle{MLM Loss.} The MLM loss predicts masked tokens in a sequence and is used widely for self-supervised learning in natural language processing (NLP)~\cite{bert}. As trajectories are sequences, so we can use the MLM loss. In particular, we apply the mask operation to a road trajectory by randomly removing some road segments. One issue is that if we mask each road segment independently, the model can easily infer masked segments because it needs to connect with its previous and subsequent segments. To avoid this issue, we mask continuous road segments with a mask length $l \geq 2$. We use grid trajectories to help reconstruct masked road trajectories, which encourages grid trajectories to provide information that complements masked road trajectories. This also makes our MLM loss different from the MLM loss in NLP and other MLM-based TRL methods~\cite{start,jgrm} in that these methods use the masked sequences themselves for reconstruction while we introduce other sequences (i.e., grid trajectories) to facilitate reconstruction. Thus, MLM loss in our model not only provides the self-supervised information but also contributes to the fusion of two modal representations.

We do not use road trajectories to help reconstruct grid trajectories. This is because grids are in a discrete free space and thus are difficult to recover, making such learning difficult. In contrast, road trajectories are continuous in the road network space, and the transitions between road segments resemble the transition between words in sentences.
A road segment can pass several grids but a grid trajectory may not contain all of them (due to GPS sampling rate, some grids of these may have no GPS point). E.g., a road trajectory is $[R_1, R_2, R_3]$, $R_1$ passes grids $[G_1, G_2]$, $R_2$ passes $[G_3, G_4, G_5]$, $R_3$ passes $[G_6, G_7, G_8]$, and the grid trajectory can be $[G_2, G_5, G_6, G_8]$. After masking (denoted as $M$), the grid trajectory is $[G_2, M_1, M_2, G_8]$, and it's difficult to predict missing grids (e.g., candidates for $M_1$ are $G_3$, $G_4$, $G_5$ of $R_2$, and $G_6$, $G_7$, $G_8$ of $R_3$). Conversely, when reconstructing masked road trajectory $[R_1, M, R_3]$, it's easy to find that $M$ is $R_2$ because $R_2$ connects with adjacent roads, and $G_5$ of this grid trajectory is on $R_2$.
The experimental results of Figure~\ref{fig:mask_recover} show that the accuracy of mask recovery of masked road trajectory reconstruction (blue line) helped by grid trajectories is much higher than that of masked grid trajectory reconstruction (red line) helped by road trajectories.

To compute MLM loss, we first use a linear layer to transform the final trajectory embedding fused by the dual-modal interactor into the predicted values for the masked road segments:
\begin{equation}
    \hat{\mathbf{Z}}^m = \textrm{Linear}(\mathbf{Z}^m) \in \mathbb{R}^{\vert T^r \vert \times \vert V \vert },
\end{equation}
Then, we use cross-entropy to compute the MLM loss for $T^r$:
\begin{equation}
    \mathcal{L}^{mlm}_{T^r} = 
     -\frac{1}{\vert T^r \vert} \sum_{\tau_i^r \in T^r} \log \frac{ \exp (\hat{\mathbf{Z}}_{\tau_i^r}^m) }{ \sum_{j = 1}^{\vert V \vert} \exp (\hat{\mathbf{Z}}_{\tau_j^r}^m) },
\end{equation}
We average the loss over the $N$ trajectories in a mini-batch to obtain the MLM loss $\mathcal{L}^{mlm}$. The overall loss in a batch is computed as:
\begin{equation}
    \mathcal{L} = \mathcal{L}^{cl} + \mathcal{L}^{mlm}.
\end{equation}

\subsection{Training and Inference}
The contrastive loss and MLM loss act on different modules in GREEN, and thus GREEN has different processing flows during training and inference.

\stitle{Training.} Since the mask operation makes road trajectories incomplete, masked trajectories cannot be used to effectively compute the contrastive loss. To address this issue, we use a \textit{two forward} operation for the road encoder during training. In particular, we first feed grid trajectories and the complete road trajectories to the grid encoder and the road encoder in the first-forward process to compute the contrastive loss. Then, we feed masked road trajectories to the road encoder for the second-forward process and feed the grid embeddings and masked road embeddings to the dual-modal interactor to compute the MLM loss. Overall, GREEN executes the grid encoder once and the road encoder twice.

\stitle{Inference.} We feed grid trajectories and the complete road trajectories as the inputs to the encoders, and the outputs from the encoders are used as the inputs to the dual-modal interactor to obtain the final trajectory representation. During inference, GREEN executes the grid encoder and the road encoder once.

We show that GREEN is comparable to state-of-the-art TRL methods in terms of training and inference efficiency in Section~\ref{sec:ablation_study}. 

 \begin{table}[tbp]
	\centering
	\caption{Statistics of the experiment datasets.}
	\label{tab:dataset}
	\resizebox{0.8\linewidth}{!}{
		\begin{tabular}{l|c|c}\toprule
			\textbf{Dataset} & Porto & Chengdu \\
			\midrule
			\textbf{\#Trajectories} & 980,644 & 1,162,303   \\
			\textbf{\#Road segments} & 10,537 & 39,503   \\
			\textbf{\#Grid cells} & 5,928 & 64,944   \\
			\textbf{Avg. traj. length(m)} & 3,196 & 3,692   \\
			% 3196.0371600502313
			% 3692.9114189607135
			\textbf{Start time} & 2013/07/01 & 2014/08/03 \\
			\textbf{End time} & 2014/06/30 & 2014/08/05 \\
			\bottomrule
		\end{tabular}
	}
\end{table}

 \begin{table*}[!t]
	\centering
	\caption{Accuracy of GREEN and the baselines for the 3 downstream tasks on Porto dataset. The best-performing baseline is marked with underline, and the bottom row is the relative improvement of GREEN over the best baseline.}
	\resizebox{0.8\linewidth}{!}{
		\begin{tabular}{l|ccc|cc|ccc}
			\toprule
			& \multicolumn{3}{c|}{Travel Time Estimation} 
			& \multicolumn{2}{c|}{Trajectory Classification} 
			& \multicolumn{3}{c}{Most Similar Trajectory Search}
			\\
			& MAE$\downarrow$  & MAPE$\downarrow$   & RMSE$\downarrow$  
			& \ \ \ Micro-F1$\uparrow$  & Macro-F1$\uparrow$  
			& \ \ MR$\downarrow$   & \ \ \ HR@1$\uparrow$     & HR@5$\uparrow$ 
			\\
			\midrule
			Traj2vec 
			& 3.816  & 0.512  & 5.239     
			& 0.515  & 0.323  
			& \ \ 60.86 & 0.326 & 0.489
			\\   
			TrajCL 
			& 1.780  & 0.189  & 2.987     
			& 0.801  & 0.761  
			& \ \ 21.93  & 0.488 & 0.643     
			\\
			Trembr 
			& 1.749  & 0.184  & 2.880     
			& 0.771  & 0.657  
			& \ \ 9.490  & 0.560  & 0.786
			\\
			PIM 
			& 2.610  & 0.361  & 4.538      
			& 0.728  & 0.613  
			& \ \ 15.35  & 0.431  & 0.612
			\\
			JCLRNT 
			& 2.331  & 0.256  & 3.778     
			& 0.782  & 0.696  
			& \ \ 12.26  & 0.472  & 0.667
			
			\\
			START 
			& \underline{1.541}  & \underline{0.173}  & \underline{2.736}     
			& \underline{0.813}  & \underline{0.772}  
			& \ \ 5.254 & 0.633  & 0.835    
			\\
			JGRM 
			& 1.649  & 0.180  & 2.803    
			& 0.805 & 0.768 
			& \ \ \underline{3.243} & \underline{0.720}  & \underline{0.904}
			\\
			\midrule
			\textbf{GREEN} 
			& \textbf{1.352}  & \textbf{0.143}  & \textbf{2.446}     
			& \textbf{0.835}  & \textbf{0.796} 
			& \ \ \textbf{1.719}  & \textbf{0.805}  & \textbf{0.968}
			\\
			\midrule
			\textbf{Imp.}
			& 12.26\%  & 17.34\%  & 10.60\%     
			& 2.71\%  & 3.11\%  
			& \ \ 46.99\% & 11.81\%  & 7.08\%
			\\
			\bottomrule
		\end{tabular}
	}
	\label{tab:performance_on_porto}
    
\end{table*}

\begin{table*}[!t]
	\centering
	\caption{Accuracy of GREEN and the baselines for the 3 downstream tasks on Chengdu dataset. The best-performing baseline is marked with underline, and the bottom row is the relative improvement of GREEN over the best baseline.}
	\resizebox{0.8\linewidth}{!}{
		\begin{tabular}{l|ccc|ccc|ccc}
			\toprule
			& \multicolumn{3}{c|}{Travel Time Estimation} 
			& \multicolumn{3}{c|}{Trajectory Classification} 
			& \multicolumn{3}{c}{Most Similar Trajectory Search}
			\\
			& MAE$\downarrow$  & MAPE$\downarrow$       & RMSE$\downarrow$  
			& \ \ F1$\uparrow$     & Accuracy$\uparrow$    & Precision$\uparrow$  
			& \ \ MR$\downarrow$   & \ \ \  HR@1$\uparrow$         & HR@5$\uparrow$ 
			\\
			\midrule
			Traj2vec 
			& 2.481  & 0.253  & 3.802   
			& \ \ 0.809  & 0.680  & 0.692
			& \ \ 51.89  & 0.415  & 0.531
			\\
			TrajCL 
			& 1.745  & 0.241  & 2.535
			& \ \ 0.854  & 0.792  & 0.816
			& \ \ 17.37  & 0.563  &  0.686           
			\\
			Trembr 
			& 1.624  & 0.207  & 2.193   
			& \ \ 0.820  & 0.723  & 0.731
			& \ \ 6.031  & 0.744  & 0.863
			\\
			PIM 
			& 1.903  & 0.248  & 3.176     
			& \ \ 0.811  & 0.704  & 0.716
			& \ \ 9.416  & 0.632  & 0.788
			\\
			JCLRNT  
			& 1.688  & 0.213  &  2.387    
			& \ \ 0.837  & 0.740  & 0.759
			& \ \ 8.529  & 0.681  & 0.790
			
			\\
			START 
			& \underline{1.351}  & \underline{0.170}       & \underline{1.862}  
			& \ \ \underline{0.871}  & \underline{0.817}  & \underline{0.840}
			& \ \ 4.852 & 0.805 & 0.912 
			\\
			JGRM 
			& 1.547  & 0.199  & 2.019     
			& \ \ 0.845  & 0.788  & 0.797
			& \ \ \underline{2.315} & \underline{0.868}  & \underline{0.926}
			\\
			\midrule
			\textbf{GREEN} 
			& \textbf{0.989}  & \textbf{0.128}  & \textbf{1.385}     
			& \ \ \textbf{0.882}  & \textbf{0.835}  & \textbf{0.855}
			& \ \ \textbf{1.026}  & \textbf{0.989}  & \textbf{0.999}
			\\
			\midrule
			\textbf{Imp.}
			& 26.79\%  & 24.71\%  &  25.62\%    
			& \ \ 1.26\%  & 2.20\%  & 1.79\%
			& \ \ 55.68\%  & 13.94\%  & 7.88\%
			\\
			\bottomrule
		\end{tabular}
	}
	\label{tab:performance_on_chengdu}
\end{table*}

\section{Experimental Evaluation} \label{section:exp}

We experiment extensively to answer the following questions:
% \squishlist
\begin{itemize}[leftmargin=*]
    \item \textbf{RQ1}: How does GREEN's accuracy compare with state-of-the-art TRL methods for various downstream tasks? 
    \item \textbf{RQ2}: How do GREEN's designs contribute to model accuracy? 
    \item \textbf{RQ3}: How effective is GREEN's pre-trained model compared with training a model from scratch for each downstream task? 
    \item \textbf{RQ4}: How efficient is GREEN at training and inference?
    \item \textbf{RQ5}: How effective is GREEN on transferability?
\end{itemize}
% \squishend

\subsection{Experiment Settings}
\noindent{\bf Datasets.} We experiment on two real-world trajectory datasets, i.e.,  Porto\footnote{https://www.kaggle.com/c/pkdd-15-predict-taxi-service-trajectory-i} and Chengdu\footnote{https://www.pkbigdata.com/common/zhzgbCmptDetails.html}, which are widely used by TRL studies~\cite{trembr,trajcl,start}. The road networks of the two cities are downloaded using OSMNX~\cite{osmnx} from OpenStreetMap\footnote{https://www.openstreetmap.org}. The proportions of training, validation, and testing data are set to [0.6, 0.2, 0.2] for both datasets. The statistics of the datasets are shown in Table~\ref{tab:dataset}. 
Road network data includes road IDs, road lengths, road types, maximum speed limit, average travel time, road direction, out-degree, in-degree, etc. Porto and Chengdu have 10,537 and 39,503 roads in the road network, respectively. Trajectory points include latitudes, longitudes, and timestamps, which are sampled every 15 seconds and 30 seconds in Porto and Chengdu on average, respectively. Also, Porto records 3 different travel modes for vehicles: 1) The trip is dispatched from the central; 2) The trip is demanded directly to a taxi driver on a specific stand; 3) Otherwise (e.g., a trip demanded on a random street). Chengdu includes whether a taxi was carrying a passenger. We remove trajectories less than 1km in length. We apply map matching~\cite{fmm} to get road trajectories. We divide the map into grids of the same size and map the GPS points to the corresponding grids to get grid trajectories. Finally, we get 980,644 and 1,162,303 trajectories in Porto and Chengdu, respectively. The average trajectory lengths are 3,196 meters and 3,692 meters in Porto and Chengdu, respectively.

% and more details are provided in Appendix~\ref{sec:appendix_datasets}.

\stitle{Baselines.} 
We compare GREEN with 7 state-of-the-art TRL methods, including 1 method that uses GPS trajectory, i.e., \textit{Traj2vec}~\cite{traj2vec}; 1 method that uses grid trajectory, i.e., \textit{TrajCL}~\cite{trajcl}, and 5 methods that use road trajectory, i.e.,  \textit{Trembr}~\cite{trembr}, \textit{PIM}~\cite{pim}, \textit{JCLRNT}~\cite{JCLRNT}, \textit{START}~\cite{start}, and \textit{JGRM}~\cite{jgrm}. Among the baselines, TrajCL is the best-performing among grid trajectory methods~\cite{trajectory_survey} while START and JGRM are the best-performing among road trajectory methods~\cite{trajectory_survey}. We choose more road trajectory methods because they are observed to achieve higher accuracy in TRL~\cite{trajectory_survey}. More details about the baselines are provided in Appendix~\ref{sec:appendix_baselines}. The implementation details of GREEN are provided in Appendix~\ref{sec:appendix_implementation}.

 \begin{figure*}[!t]
	\centering
	\includegraphics[width=\linewidth]{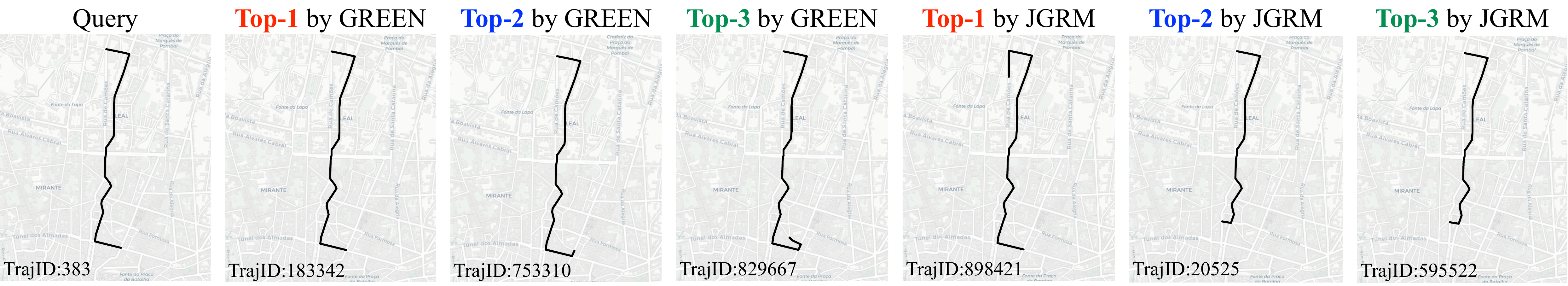}
	\caption{Comparison of the top-3 similar trajectories retrieved by GREEN and JGRM from Porto dataset for a query trajectory.}
	\label{fig:exp_show_sim}
\end{figure*}

\stitle{Accuracy Measures.}
We conduct three downstream tasks introduced in Section~\ref{subsec:problem}, i.e., travel time estimation, trajectory classification, and most similar trajectory search, which are widely used to evaluate TRL methods~\cite{JCLRNT,start,jgrm}. Among these tasks, travel time estimation and trajectory classification fine-tune the model using data, while most similar trajectory search directly uses the learned trajectory representations to compute similarity. The settings of downstream tasks are described in Appendix~\ref{sec:appendix_downstream settings}. We also follow existing works to choose accuracy measures for these tasks. In particular, for travel time estimation, we report the mean absolute error (MAE), mean absolute percentage error (MAPE), and root mean square error (RMSE). For multi-class trajectory classification on the Porto dataset, we report Micro-F1 and Macro-F1. For binary trajectory classification on the Chengdu dataset, we report the F1-score, accuracy, and precision. For most similar trajectory search, we report the hit ratio at the top-k results, i.e., the HR@1, HR@5, and the mean rank (MR). Ideally, the MR value is 1.

\subsection{Main Results (RQ1)}
Table~\ref{tab:performance_on_porto} and Table~\ref{tab:performance_on_chengdu} compare the accuracy of GREEN with the 7 baselines. We make several observations as follows.

First, GREEN consistently performs the best among the methods for all three downstream tasks on two experiment datasets. This asserts the effectiveness and generality of GREEN. Moreover, the improvement of GREEN over the best-performing baseline is usually significant. For travel time estimation, the improvement is  10.60\% at the minimum and can be over 26\% at the maximum. The case is similar for most similar trajectory search, with the minimum improvement at 7.08\% and the maximum improvement above 55\%. This is because travel time estimation benefits from the timing information from both grid and road trajectories, and most similar trajectory search benefits from expressing a trajectory in both grid and road forms. The improvement of GREEN for trajectory classification is smaller than the other tasks because classification is relatively easy, especially for binary classification on Chengdu, which is evidenced by the high accuracy of the baselines.

Second, road trajectory methods generally perform better than grid trajectory methods. Both START and JGRM use road trajectory and are the best-performing baselines. This is because road trajectory provides more travel information by constraining the movement to be on continuous road segments while grid trajectory consists of discrete grids. This is in line with the trend that most recent TRL methods use road trajectory. Moreover, it also supports our decision to use the grid trajectory to help reconstruct the road trajectory in the MLM loss, i.e., the road trajectory has more regularity and thus is easier to reconstruct.   

Third, contrastive learning methods (e.g., JGRM and START) generally perform better than auto-regressive methods (e.g., Trembr and Traj2vec). This is because the auto-regressive methods are trained by generating each trajectory incrementally in steps like the next token prediction in NLP. The lengths of the trajectories are skewed, and the accumulated errors lead to poor performance for long trajectories. Contrastive learning uses data augmentations to introduce more supervision signals for training so that the learned representations exhibit better generalization. This observation suggests that TRL needs good objectives for supervision signals and motivates our contrastive and MLM loss.   

To provide an intuitive illustration of the effectiveness of GREEN, Figure~\ref{fig:exp_show_sim} compares the most similar trajectories returned by it and JGRM, the best-performing baseline for trajectory similarity search. We observe that trajectories returned by GREEN are more similar to the query trajectory than the  trajectories returned by JGRM.

\begin{table}[!t]
   \LARGE
   \centering
   \caption{Accuracy of GREEN when disabling the key designs. We denote travel time estimation as TTE, trajectory classification as TC, and most similar trajectory search as MSTS.}
   \resizebox{\linewidth}{!}{
   \begin{tabular}{l|ccc|ccc}\toprule
      & \multicolumn{3}{c}{Porto} & \multicolumn{3}{c}{Chengdu}
      \\\cmidrule(lr){2-4}\cmidrule(lr){5-7}  
        & MAE$\downarrow$  & Micro-F1$\uparrow$ & MR$\downarrow$ 
        & MAE$\downarrow$  & F1$\uparrow$ & MR$\downarrow$   \\
        & (TTE)  & (TC) & (MSTS) 
        & (TTE)  & (TC) & (MSTS)   \\
        \midrule
	w/o Grid enc.
        & 1.866  & 0.826  & 17.79
        & 1.240  & 0.870  & 10.51
        \\
	w/o Road enc. 
 	& 1.447 & 0.830 & 28.58
        & 1.108 & 0.880 & 17.43
        \\
        w/o CL loss 
	& 1.386  & 0.833  & 12.74 
 	& 1.014  & 0.876  & 4.629 
        \\
        w/o MLM loss
	& 1.492  & 0.819  & 2.303  
 	& 1.062  & 0.851  & 1.609  
        \\
	w/o 2-forward
	& 1.431 & 0.830 & 1.801  
 	& 1.016 & 0.870 & 1.257 
        \\
	\midrule
	\textbf{GREEN} 
        & \textbf{1.352} & \textbf{0.835} & \textbf{1.719} 
        & \textbf{0.989}  & \textbf{0.882}  & \textbf{1.026} 
        \\
	\bottomrule
   \end{tabular}
   }
   \label{tab:ablation}
\end{table}

\subsection{Micro Experiments} \label{sec:ablation_study}

\noindent{\bf Ablation Study (RQ2).} To validate GREEN's designs, we experiment with 5 of its variants:
% \squishlist
\begin{itemize}[leftmargin=*]
    \item \textbf{w/o Grid enc.} keeps the road encoder and uses the MLM loss.
    \item \textbf{w/o Road enc.} keeps the grid encoder and uses the MLM loss.
    \item \textbf{w/o CL loss} removes the contrastive loss.
    \item \textbf{w/o MLM loss} removes the MLM loss. As MLM loss acts on the dual-modal interactor, we also remove the interactor and compute the final trajectory representation by directly averaging the outputs of the road and grid encoders.
    \item \textbf{w/o 2-forward} removes the two forward operations in training, and the road encoder uses masked road trajectories as inputs.
\end{itemize}
% \squishend

Table~\ref{tab:ablation} reports the results of the ablation study. 
For brevity,
we use only one accuracy measure for each task. The results suggest that all our designs are effective in improving model accuracy since removing any of them degrades performance. However, the contributions of the designs vary for different tasks. In particular, the MLM loss is more important for trajectory classification than the contrastive loss while the opposite is true for travel time estimation. Moreover, using only the grid encoder leads to better performance than the road encoder for both trajectory classification and travel time estimation while the opposite holds for most similar trajectory search. Compared with the other two tasks, most similar trajectory search has larger degradation when disabling the designs. This is because it directly uses the trajectory representations for similarity computation while the other tasks fine-tune the model.

\begin{figure}[!t]
    \centering
    \includegraphics[width=\linewidth]{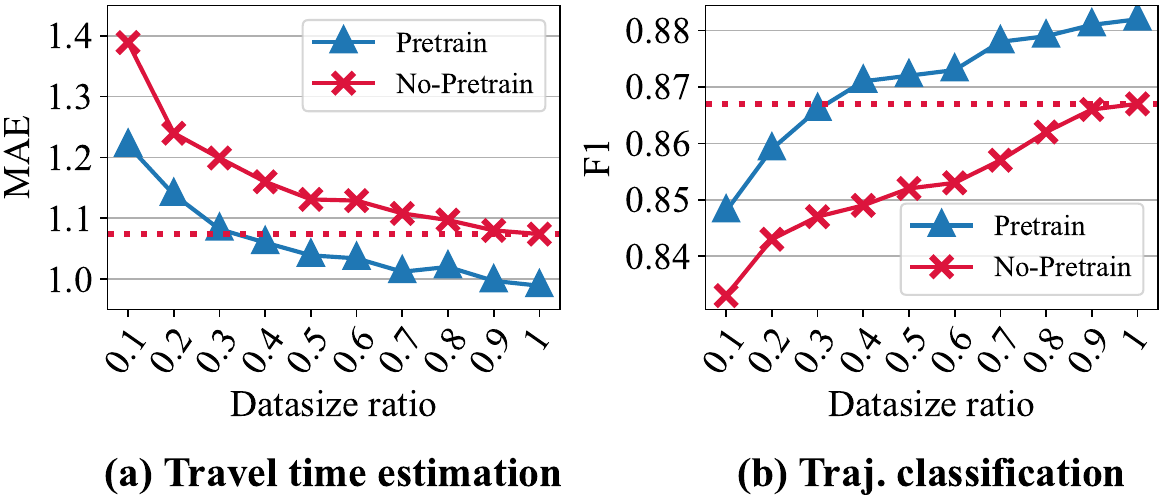}
    \caption{The effect of GREEN pre-training on Chengdu. X-axis is the ratio of the data used for training in the dataset.}
    \label{fig:exp_pretraining}
\end{figure}

\stitle{Effect of Pre-training (RQ3).}
As mentioned above, travel time estimation and trajectory classification fine-tune the model produced by GREEN. As such, GREEN can be viewed as a pre-training for the two tasks. To validate the effectiveness of GREEN, we compare it with a baseline that directly trains the models for the two tasks (i.e., \textit{No-Pretrain}). We ensure that \textit{No-Pretrain} adopts the same model architecture as GREEN for a fair comparison but the loss terms are changed to cross-entropy for trajectory classification and squared error for travel time estimation. Figure~\ref{fig:exp_pretraining} reports the results when changing the portion of used training data from 10\% to 100\%. We observe that both \textit{Pretrain} and \textit{No-Pretrain} improve accuracy when using more training data but \textit{Pretrain} achieves much higher accuracy than \textit{No-Pretrain} when using the same amount of training data. Notably, \textit{Pretrain} matches the accuracy of \textit{No-Pretrain} on the full dataset when using only 30\% of the training data. This suggests that our contrastive loss and MLM loss are effective, and they help GREEN to learn information that benefits the downstream tasks.  

\begin{figure}[!t]
    \centering
    \includegraphics[width=\linewidth]{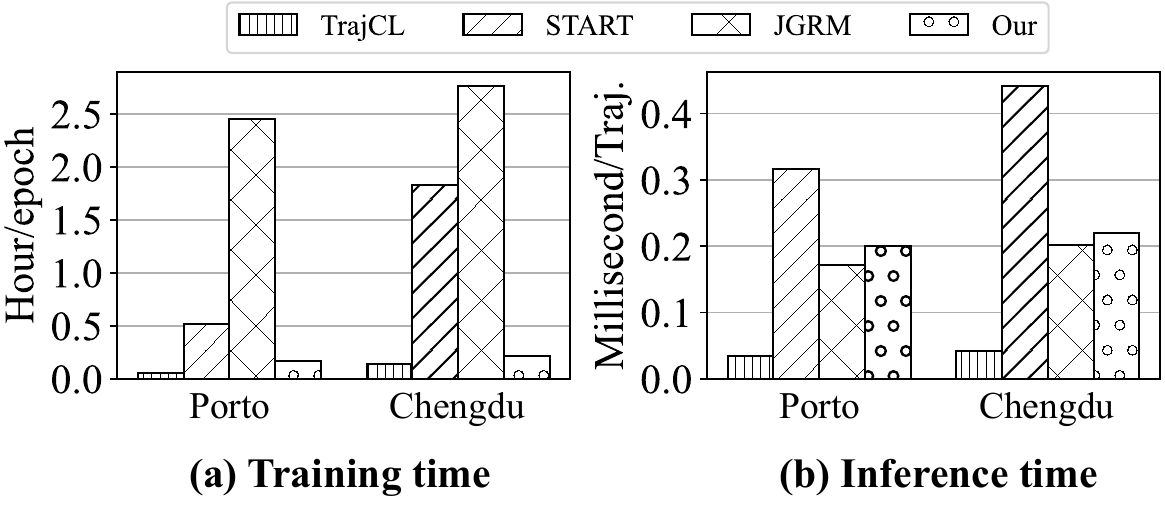}
    \caption{Model training time (in hours) for one epoch and model inference time (in ms) of one trajectory.}
    \label{fig:exp_time}
\end{figure}

\stitle{Efficiency Study (RQ4).} In Figure~\ref{fig:exp_time}, we compare the training and inference time of GREEN with TrajCL, START, and JGRM. TrajCL represents grid trajectory methods while START and JGRM represent road trajectory methods and the best-performing baselines. For training, we report the time for one epoch, while for inference, we report the time to compute the representation for a trajectory. Preprocessing time, such as map matching, is not included here.

The results show that GREEN has much shorter training time than START and JGRM, and its inference time is similar to JGRM. Thus, compared with state-of-the-art baselines, we do not have to pay extra computation costs for GREEN's improved accuracy. TrajCL is fast for both training and inference because grid trajectories are generally shorter in length than road trajectories and its model is simple. JGRM is slow in training but becomes faster in inference because it uses many loops to align GPS points with road segments during training (JGRM uses an assignment matrix to save alignment information during preprocessing, which is further used for model training) while inference does not conduct the alignments. START has long training and inference time because it uses data augmentations in training and its model is complex. Moreover, our model also converges quickly and the loss stabilizes after 10 epochs, and for START and JGRM, the loss stabilizes after 15 epochs.

\begin{table}[!t]
   \LARGE
   \centering
   \caption{Accuracy of GREEN on transferability between Porto and Chengdu.}
   
   \resizebox{\linewidth}{!}{
   \begin{tabular}{l|ccc|ccc}\toprule
       
      \multirow{2}*{\textbf{Porto->Chengdu}} 
      & \multicolumn{3}{c}{Travel Time Estimation} & \multicolumn{3}{c}{Trajectory classification} \\
      \cmidrule(lr){2-4}\cmidrule(lr){5-7}  
      & MAE$\downarrow$  & MAPE$\downarrow$ & RMSE$\downarrow$ 
      & F1$\uparrow$  & Accuracy$\uparrow$ & Precision$\uparrow$   \\
      \midrule
      w/ \ \ Transfer
      & 1.046  & 0.138  & 1.465
      & 0.833  & 0.764  & 0.802
      \\
      w/o Transfer
      & 0.989  & 0.128  & 1.385
      & 0.882  & 0.835  & 0.855
      \\
      \bottomrule
   \end{tabular}
   }
   \vspace{1.5mm}

   \resizebox{\linewidth}{!}{
   \begin{tabular}{l|ccc|cc}\toprule
      \multirow{2}*{\textbf{Chengdu->Porto}} & \multicolumn{3}{c}{Travel Time Estimation} & \multicolumn{2}{c}{Trajectory classification} \\
      \cmidrule(lr){2-4}\cmidrule(lr){5-6}  
      & MAE$\downarrow$  & MAPE$\downarrow$ & RMSE$\downarrow$ 
      & Micro-F1$\uparrow$  & Macro-F1$\uparrow$  \\
      \midrule
      w/ \ \ Transfer
      & 1.429  & 0.157  & 2.497
      & 0.823  & 0.781  
      \\
      w/o Transfer
      & 1.352 & 0.143   & 2.446 
      & 0.835  &  0.796 
      \\
      \bottomrule
   \end{tabular}
   }

   \label{tab:transferability}
\end{table}

\stitle{Transferability (RQ5).} The transferability of the pre-trained model shows the reuse ability. Here, we study the transferability of our model between different cities. Specifically, we first pre-train GREEN on City A, then fine-tune it on City B for a task (\textbf{City A->City B}). Due to the different urban structures, we randomly initialize the CNN and GNN modules during transferring and keep the rest parameters of the pre-trained model. We do not evaluate the similarity performance because it is not a fine-tuning task and re-initializing parameters results in significant performance degradation. 
The results in Table~\ref{tab:transferability} show that although there is a performance degradation compared to no transferring (pre-training and fine-tuning both on City B), it still outperforms most baselines. Porto->Chengdu has more performance degradation because the data scale of Chengdu is larger, and more token information is introduced during transferring, i.e., grid IDs and road segment IDs.

\stitle{Hyper-parameters.} We also explore how the hyper-parameters (e.g., grid cell size, embedding dimension, continuous mask ratio and length, and network layers) affect the performance of GREEN. Due to the page limit, we report the results in Appendix~\ref{sec:appendix_hyper_param}.

\section{Conclusion}
We propose GREEN, a self-supervised method that utilizes multimodal learning
for trajectory representation learning (TRL). GREEN exploits that GPS trajectories can be expressed as both grid and road trajectories that complement each other by capturing different information. To utilize the two trajectory expressions jointly, we tailor an encoder for each expression that captures their respective information and carefully design loss terms to align the two encoders and fuse their outputs into a final trajectory representation. Experiment results show that
GREEN is capable of consistently outperforming existing TRL methods, often by a large margin.

\section{Acknowledgement}
This paper was supported by the National Key R\&D Program of China 2024YFE0111800, and NSFC U22B2037, U21B2046, and 62032001.

\bibliographystyle{ACM-Reference-Format}
\bibliography{green}

\appendix
\section{Appendix}\label{sec:appendix}
\subsection{Baselines} \label{sec:appendix_baselines}
\noindent
\textbf{GPS-based trajectory representation learning:}
\squishlist
    \item \textbf{Traj2vec}\footnote{https://github.com/yaodi833/trajectory2vec}: An RNN-based seq2seq model converts GPS trajectory into feature sequence to learn trajectory representations.
\squishend

\noindent
\textbf{Grid-based trajectory representation learning:}
\squishlist
    \item \textbf{TrajCL}\footnote{https://github.com/changyanchuan/TrajCL}: The state-of-the-art grid-based method proposes a set of trajectory augmentations on grid trajectory in free space and dual-feature self-attention to learn grid trajectory representations using contrastive learning with the Transformer.
\squishend

\noindent
\textbf{Road-based trajectory representation learning:}
\squishlist
    \item \textbf{Trembr}: An RNN-based seq2seq model uses spatial-temporal road trajectory sequences to learn trajectory representation.
    \item \textbf{PIM}\footnote{https://github.com/Sean-Bin-Yang/Path-InfoMax}: It first uses node2vec~\cite{node2vec} to learn road embeddings on the road network and then proposes mutual information maximization to learn trajectory representations.
    \item \textbf{JCLRNT}\footnote{https://github.com/mzy94/JCLRNT}: It uses GAT and Transformer to learn road representations and trajectory representations. Three different contrastive losses are designed to optimize the model.
    \item \textbf{START}\footnote{https://github.com/aptx1231/START}: The state-of-the-art method on the road trajectory representation uses BERT~\cite{bert} as the trajectory encoder that integrates travel semantics and temporal regularities with contrastive learning and two self-supervised tasks.
    \item \textbf{JGRM}\footnote{https://github.com/mamazi0131/JGRM}: The state-of-the-art road-based method models trajectory representation with GPS and road information. It learns GPS points as another view representation of the road to learn road trajectory representations with three self-supervised tasks.
\squishend

\subsection{Implementation}
\label{sec:appendix_implementation}
By default, the grid cell size is set to 100m$\times$100m, the convolution kernel size is 3$\times$3, and the output channel $C^\prime = 64$ for the grid encoder. We set the dimension of the trajectory representation $d = 128$, the hidden embedding size $h = 2d$, the number of GAT layers to 3, the number of the Transformer layers in road encoder to 4, the number of the Transformer layers in grid encoder to 2, and the number of dual-modal interactor layers to 2. We use 4 attention heads for the GAT, the Transformer both in encoders, and 2 attention heads for the dual-modal interactor. The dropout ratio is 0.1. The mask ratio is set as 0.2 and the continuous mask length is set as 2 for the MLM loss. The temperature $\delta$ of the contrastive loss is set to 0.07.
We pre-train GREEN using the Adam~\cite{adam} optimizer. The batch size is 128, the number of training epochs is 30, and the learning rate is 2e-4 for both datasets.

\subsection{Downstream Task Settings} \label{sec:appendix_downstream settings}
\noindent
\textbf{Travel Time Estimation.} Travel time estimation is a fine-tuning task, which aims to predict the travel time of a trajectory. We add a two-layer MLP after the dual-modal interactor to predict travel time on \textit{minutes} and use mean square error as a loss function. Since inputs of the pre-training model have full-time information, we only retain the start time and ignore other times to avoid information leakage in fine-tuning. We set the learning rate to 1e-4 and other hyper-parameters to be consistent with pre-training.

\stitle{Trajectory Classification.} Trajectory classification is also a fine-tuning task, which assigns a trajectory to a category. We classify trajectories according to the travel mode on Porto, which is a multi-classification. We determine if there are passengers in the taxi of Chengdu, which is a binary-classification. We add a linear layer after the dual-modal interactor to predict the category label and use cross-entropy as a loss function. We set the learning rate to 1e-4 and other hyper-parameters to be consistent with pre-training.

\stitle{Most Similar Trajectory Search.} Given a query trajectory $T$ of the query dataset $\mathcal{D}_q$, the most similar trajectory search is to find the most similar trajectory $T^\prime$ from a large trajectory database $\mathcal{D}_d$. In this task, we directly use pre-trained trajectory representations to evaluate the performance. However, the lack of ground truth makes it difficult to evaluate the accuracy. TrajCL~\cite{trajcl} proposes a downsample method for grid trajectory to obtain sub-grid trajectories as ground truths and JCLRNT~\cite{JCLRNT} proposes a detour method for road trajectory to get the most similar trajectory. However, \textit{their methods can not be applied to both the grid trajectory and road trajectory}. To solve this problem, we first keep the start and end GPS points unchanged to keep their overall origin-destination (OD) similar, then downsample other GPS points to obtain sub-grid trajectories, and also perform map matching to obtain new road trajectories. However, the influence of downsampling can be reduced by map matching on road trajectories. To eliminate this influence, we first use a high downsampling ratio, i.e., 0.5, then we choose the road trajectory in the new road trajectories with a change rate between 0.3 and 0.5, the corresponding sub-grid trajectories as the positive set $\mathcal{D}_p$, and the corresponding raw road and grid trajectories as the query set $\mathcal{D}_q$. Specifically, we first randomly select 1,000 trajectories as the query $\mathcal{D}_q^\prime \in \mathcal{D}_q$ and the corresponding positive trajectories $\mathcal{D}_p^\prime \in \mathcal{D}_p$ as label trajectories. Then we randomly select other 100,000 trajectories as negative trajectories set $\mathcal{D}_n$ from the dataset. Together, we get the final database $\mathcal{D}_d = \mathcal{D}_p^\prime \cup \mathcal{D}_n$.

\begin{table}[!t]
   \centering
   \caption{Effect of grid cell size in Porto. The best-performing and the second-performing are marked with bold and underline, respectively.}
   \resizebox{0.8\linewidth}{!}{
   \begin{tabular}{c|ccc}
        \toprule
        & MAE$\downarrow$  & Micro-F1$\uparrow$ & MR$\downarrow$ \\
        & (TTE)  & (TC) & (MSTS)   \\
        \midrule
        50m $\times$ 50m
	& \textbf{1.130}  & \underline{0.834}  & 1.856
        \\
        100m $\times$ 100m
	& \underline{1.352}  & \textbf{0.835}  & \underline{1.719}    
        \\
	200m $\times$ 200m
	& 1.541 & 0.833 & \textbf{1.602}    
        \\
	\bottomrule
   \end{tabular}
   }
   \label{tab:grid_size}
\end{table}

\subsection{The Effect of Hyper-parameters}\label{sec:appendix_hyper_param}

\stitle{Grid cell size.} We vary the grid cell size of grid trajectories among [50, 100, 200]. The results of Table~\ref{tab:grid_size} show that a larger grid degrades travel time estimation (TTE), improves most similar trajectory search (MSTS), and does not affect trajectory classification (TC) much. This is because a large grid size reduces the spatial irregularity of GPS points and improves region information, which benefits MSTS. However, time information is reduced when more GPS points are gathered in a grid cell, which degrades TTE. TC relies on the travel semantics of trajectories and thus is not affected when the grid size is in a reasonable range. 100m is the default grid size because it balances among different tasks.

\begin{figure}[!t]
	\centering
	\includegraphics[width=\linewidth]{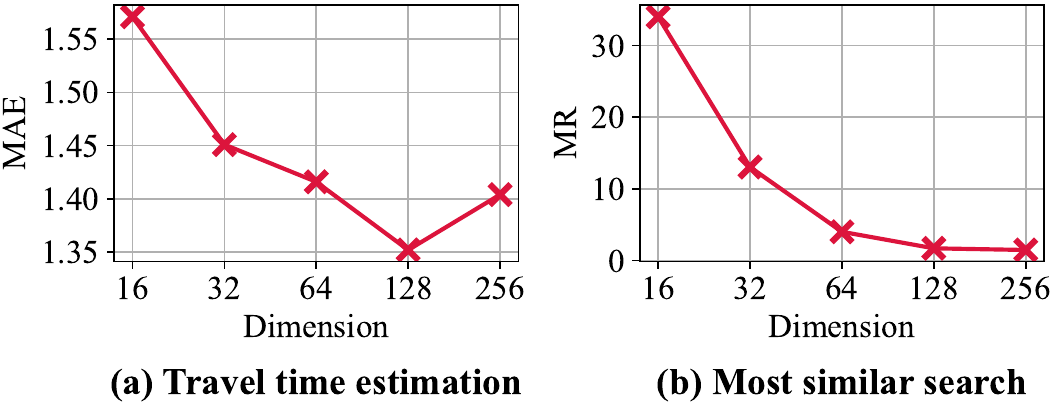}
	\caption{Effect of embedding dimension in Porto.}
	\label{fig:exp_dimension}
\end{figure}

\stitle{Dimension.}
We vary the dimension of the final trajectory representation among [16, 32, 64, 128, 256] and keep the hidden dimension as 2x of the embedding dimension. Figure~\ref{fig:exp_dimension} shows that both tasks generally improve when using a larger embedding dimension. This is because a larger embedding dimension allows to encode more information of travel semantics in the trajectory representation. The accuracy of travel time estimation drops when increasing from 128 to 256 dimension because it fine-tunes a model on the trajectory representations, and over-fitting happens when there are too many parameters at large embedding dimension.   

\balance

\begin{figure}[!t]
	\centering
	\includegraphics[width=\linewidth]{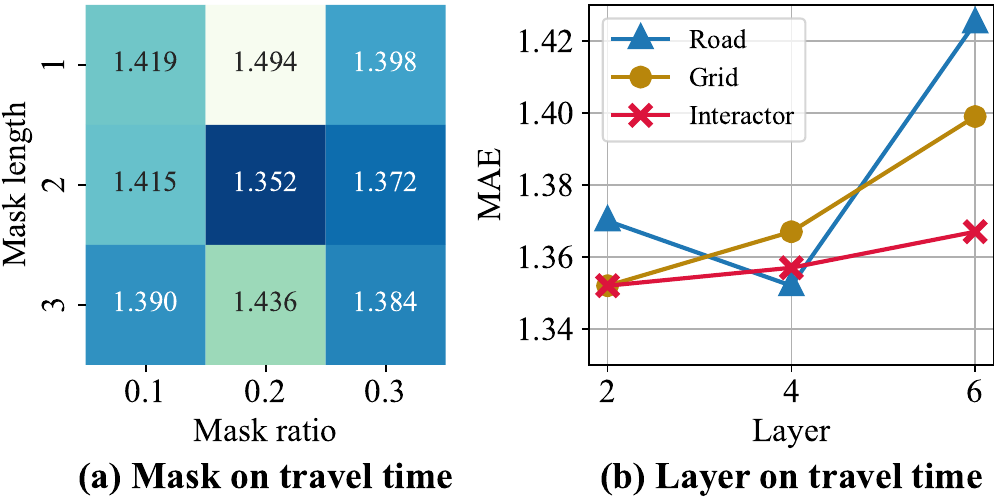}
	\caption{Effect of mask and layer in Porto.}
	\label{fig:mask_layer}
\end{figure}

\stitle{Mask.} 
We vary the mask length among [1, 2, 3] and the mask ratio among [0.1, 0.2, 0.3] when masking the road trajectory for the MLM loss. The results in Figure~\ref{fig:mask_layer} (a) show that the optimal performance is achieved at an intermediate mask length (i.e., 2) and mask ratio (i.e., 0.2). This is because small values for mask length and mask ratio make the task trajectory reconstruction easy, and the model may not learn enough information. When the values are too large, the task is too difficult, and learning is also hindered. 

\stitle{Layer.} 
We vary the number of layers for the road encoder, grid encoder, and dual-modal interactor among [2, 4, 6]. The remaining hyper-parameters are kept at default values. The results are shown in Figure~\ref{fig:mask_layer} (b). When the number of layers of road encoder, grid encoder, and dual-modal interactor are 4, 2, and 2, respectively, the performance is the best. Specifically, 
the average length of grid trajectories is smaller than road trajectories, and the travel semantics carried by grid trajectories is lower than road trajectories, so it does not need too many network layers. On the contrary, the road trajectory requires more network layers. Meanwhile, because the two encoders and contrastive loss we designed have effectively learned and aligned the trajectory representation, only a few network layers of interactor are required to achieve the desired effect.

% \end{sloppypar}
\end{document}